\documentclass[letterpaper, 10 pt, conference]{ieeeconf}  % 
\IEEEoverridecommandlockouts                              % 
\usepackage{graphicx} 
\usepackage{algorithm}
\usepackage{algorithmic}
\usepackage{lipsum} 
\usepackage{flushend}
\usepackage{float}
\usepackage{booktabs}
\usepackage{xcolor}
\usepackage{url}

\usepackage{amsmath}
\usepackage{amssymb}
\usepackage{booktabs}
\title{\LARGE \bf
Temporal Attention for Cross-View Sequential Image Localization
}
\author{Dong Yuan$^{1}$, Frederic Maire$^{1}$, Feras Dayoub$^{2}$ 
\thanks{{$^{1}$QUT Centre for Robotics, Queensland University of Technology, Australia} {\tt \emph{\{yuand2, f.maire\}@qut.edu.au}}} 
\thanks{{$^{2}$Australian Institute for Machine Learning (AIML), University of Adelaide, Australia}  {\tt \emph{feras.dayoub@adelaide.edu.au}}}
}

\begin{document}
\maketitle
\thispagestyle{empty}
\pagestyle{empty}
%%%%%%%%%%%%%%%%%%%%%%%%%%%%%%%%%%%%%%%%%%%%%%%%%%%%%%%%%%%%%%%%%%%%%%%%%%%%%%%%
\begin{abstract}
%In this paper, we introduce a novel approach to robot cross-view localization, focusing on the fine-grained localization of sequential street-view images within a single known satellite image patch. Traditionally, cross-view localization is based on one-to-one image retrieval or focuses on single street-view image fine-grained localization. We extend the cross-view localization task into a sequential street-view images fine-grained localization task, thereby rendering it more applicable and viable for real-world scenarios. To address this task, we propose a model that utilizes a novel temporal attention module to leverage information from previous images for improved sequential image localization accuracy. Our approach significantly reduces mean and median localization errors compared to current state-of-the-art methods on the single image fine-grained localization task, as demonstrated on the Cross-View Image Sequence (CVIS) dataset. By adapting the KITTI-CVL dataset into sequential image sets, we provide a more realistic and practical dataset for future research. Our results on the KITTI-CVL dataset showcase the model's robust generalization across different times and unknown areas, marking a reduction in mean distance error by 75.3\% in cross-view sequential image localization and offering promising implications for robotic navigation and mapping technologies.

This paper introduces a novel approach to enhancing cross-view localization, focusing on the fine-grained, sequential localization of street-view images within a single known satellite image patch, a significant departure from traditional one-to-one image retrieval methods. By expanding to sequential image fine-grained localization, our model, equipped with a novel Temporal Attention Module (TAM), leverages contextual information to significantly improve sequential image localization accuracy. Our method shows substantial reductions in both mean and median localization errors on the Cross-View Image Sequence (CVIS) dataset, outperforming current state-of-the-art single-image localization techniques. Additionally, by adapting the KITTI-CVL dataset into sequential image sets, we not only offer a more realistic dataset for future research but also demonstrate our model's robust generalization capabilities across varying times and areas, evidenced by a 75.3\% reduction in mean distance error in cross-view sequential image localization.
\end{abstract}

\section{Introduction}
Cross-view image geo-localization endeavours to enhance the precision of robot localization in outdoor environments that may have erratic GPS signals. Most previous works~\cite{hu2018cvm,shi2019spatial,regmi2019bridging,liu2019lending,shi2020optimal,zhu2022transgeo,shi2020looking,cai2019ground,yang2021cross} formulate the cross-view geo-localization problem as an image retrieval task, and solve it by metric learning. When provided with a current street-view image captured by the robot, they retrieve the corresponding satellite image patch from a reference dataset, assigning the center GPS coordinate of the retrieved satellite patch as the location of the street-view image. Despite achieving a commendable level of accuracy in image retrieval, these results rely on a one-to-one setting where one street-view image is exactly located at the center of one satellite image from the reference dataset. In real-world scenarios, it is impractical to guarantee that the current street-view image precisely aligns with the center of a specific satellite image within the reference dataset. Consequently, employing the center GPS coordinate of the retrieved satellite image as the current location invariably leads to estimation errors.

\begin{figure}
    \centering
    \includegraphics[width=1\linewidth]{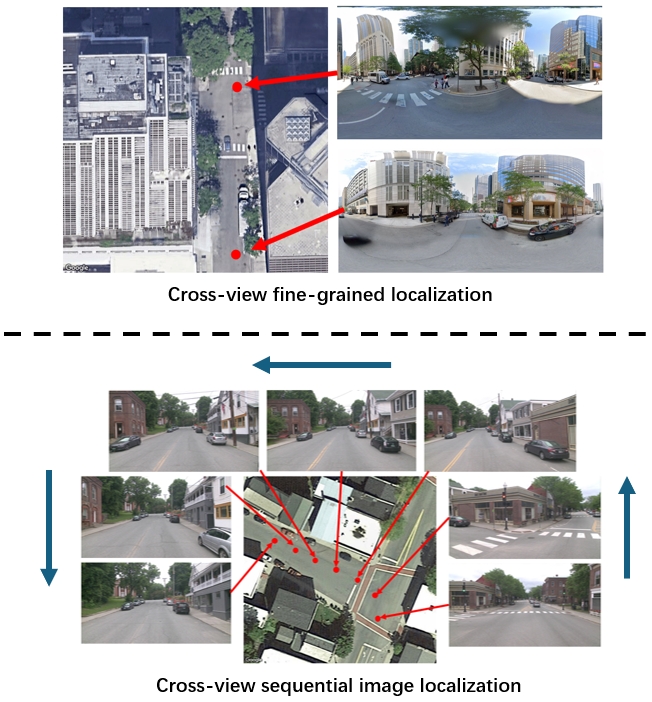}
    \caption{Comparison of two cross-view localization tasks. \textbf{Top:} Cross-view fine-grained localization primarily emphasizes discrete location predictions, where street-view images are widely spaced apart and exhibit discontinuous distribution along the route or area of interest. \textbf{Bottom:} Cross-view sequential image localization is centered around predicting the location in a satellite image of each street-view image belonging to a sequence, aligning more closely with practical localization applications.}
    \label{fig:tasks}
\end{figure}

Departing from the idea of image retrieval, recent works~\cite{yuan2024cross,xia2022visual,zhu2021vigor} focus on the cross-view fine-grained localization task, which involves predicting pixel coordinates corresponding to the current street view on a given satellite image patch (see Fig.\ref{fig:tasks}). Zhu \textit{et al.}~\cite{zhu2021vigor} introduced the VIGOR dataset for research purposes and proposed a regression-based approach to predict the offset of the street-view location relative to the center point of the satellite image. Xia \textit{et al.}~\cite{xia2022visual} employed the encoding of satellite images and street-view images into local and global descriptors, respectively. Using correlation operations between the global descriptor and each local descriptor, a dense location distribution map is derived. Yuan \textit{et al.}~\cite{yuan2024cross} utilized cross-attention mechanisms to establish correspondences between two views. They then integrated classification and regression prediction methods, resulting in state-of-the-art performance on the cross-view fine-grained localization task.

We argue that the aforementioned cross-view fine-grained localization studies predominantly concentrate on the localization of individual street-view images. However, in practical robot localization applications, continuous streams of ground-view images are typically received from the robot's cameras. Consequently, extending the cross-view fine-grained localization task to sequential ground-view image localization is more practical and more acceptable in real-world scenarios.  

In this paper, we refer to the novel task as \textit{cross-view sequential image localization}, involving the prediction of sequential ground-view image location within a known satellite image patch, as illustrated in Fig.\ref{fig:tasks}. To leverage temporal information effectively, we adopt the state-of-the-art method on the cross-view fine-grained task~\cite{yuan2024cross} as the foundational framework and augment it by adding a hidden state to retain information from the preceding time step. Subsequently, we introduce a novel temporal attention module designed to fuse information from both the hidden state and the current state, thereby generating updated features for the prediction of the current street-view location. We evaluate our method on a large-scale cross-view image sequence dataset (CVIS)~\cite{zhang2023cross}, which contains thousands of sequential ground-view images with corresponding satellite images. We also partition the KITTI-CVL dataset~\cite{shi2022cvlnet} into numerous ground-view image sequences, facilitating the assessment of our model's generalization across different times and unknown areas. Our main contributions can be summarized as follows: 
    \begin{itemize}
        % \item We propose a temporal attention mechanism that effectively leverages the temporal information present in sequential ground-view images. This mechanism integrates information from both the current and previous states, allowing for more accurate prediction of the sequential street view location relative to the corresponding satellite image.
        \item We introduce a novel temporal attention mechanism that enhances the model's ability to utilize historical context from sequential ground-view images for more accurate localization.
        \item We enhance the KITTI-CVL dataset by segmenting it into realistic short image sequences, creating a refined dataset for evaluating cross-view sequential image localization. This effort, coupled with our use of the CVIS dataset, underpins our experiments, showing our method's improved accuracy and robustness.
    \end{itemize}
To foster future research on cross-view sequential image localization, we make our code available at: \url{https://github.com/UQ-DongYuan/CVSeqLocation}

\section{Related Work}
\subsection{Cross-view Image Retrieval}
Cross-view image retrieval methodologies commonly entail the extraction of high-dimensional feature vectors, referred to as image embeddings, from ground and aerial view images. The determination of matching between images from different views is achieved through the computation of the distance between their respective image embeddings. Hence, the pivotal factor in enhancing retrieval accuracy lies in the extraction of high-quality features from ground and aerial view images. CVM-Net~\cite{hu2018cvm} employed NetVLAD~\cite{arandjelovic2016netvlad} layers to extract view-invariant features to improve performance. 
Shi \textit{et al.}~\cite{shi2020optimal} introduced a feature transport module designed to align features extracted from both view images. Bridging the domain gap between ground and aerial views can also lead to an enhancement in retrieval accuracy. Regmi \textit{et al.}~\cite{regmi2019bridging} employed a conditional Generative Adversarial Networks (cGAN)~\cite{isola2017image} to synthesize aerial images from ground-view images, subsequently extracting features from the synthesised images. SAFA~\cite{shi2019spatial}, on the other hand, leveraged prior geometric knowledge to convert aerial images into ground-view by utilizing polar transformation, and proposed a spatial-aware module to enhance the spatial correlation of extracted features. Toker \textit{et al.}~\cite{toker2021coming} combined SAFA~\cite{shi2019spatial} with a cGAN to achieve promising results. Building upon the success of Transformer~\cite{vaswani2017attention} in image recognition~\cite{dosovitskiy2020image} and object detection~\cite{carion2020end}, ~\cite{zhu2022transgeo} and ~\cite{yang2021cross}  leveraged Transformer as the backbone for feature extraction from both ground-view and aerial view images. Through the incorporation of position embedding and self-attention mechanisms~\cite{vaswani2017attention}, the extracted features are enriched with spatial information, consequently improving the accuracy of retrieval. 

While the aforementioned methods have demonstrated high retrieval accuracy, their accuracy in location estimation is limited by the assumption of center alignment and the setting of one-to-one matching.

\subsection{Cross-view Fine-grained Localization}
To address the limitations inherent in cross-view image retrieval, Zhu \textit{et al.}~\cite{zhu2021vigor} established the VIGOR dataset, wherein each street-view image is covered by multiple satellite images and positioned non-centrally within the satellite image. They explored the application of regression methods to predict the offset of street-view images relative to the center of satellite images. Building upon the work of SAFA~\cite{shi2019spatial}, Xia \textit{et al.}~\cite{xia2022visual} extracted multiple local descriptors as well as a global descriptor from satellite and street-view images, respectively. By conducting correlation operations between these descriptors, a similarity score map is generated. This map is then upsampled to obtain a location probabilistic distribution map of the same dimensions as the satellite image, facilitating the prediction of pixel locations of street-view images within satellite images. Yuan \textit{et al.}~\cite{yuan2024cross} denoted this task as cross-view fine-grained localization, introducing the cross-attention mechanism to compute the cross-attention score between extracted features from two views. Leveraging this score, they obtained fusion features that integrate correspondences between the two views. Finally, employing a combination of classification and regression akin to object detection tasks, they further reduce the location prediction errors.

Our proposed method extends the foundational framework~\cite{yuan2024cross} by introducing a temporal attention module to handle temporal information, enabling the prediction of pixel-wise locations for each street-view image within an image sequence.

\begin{figure*}
    \centering
    \includegraphics[width=1\linewidth]{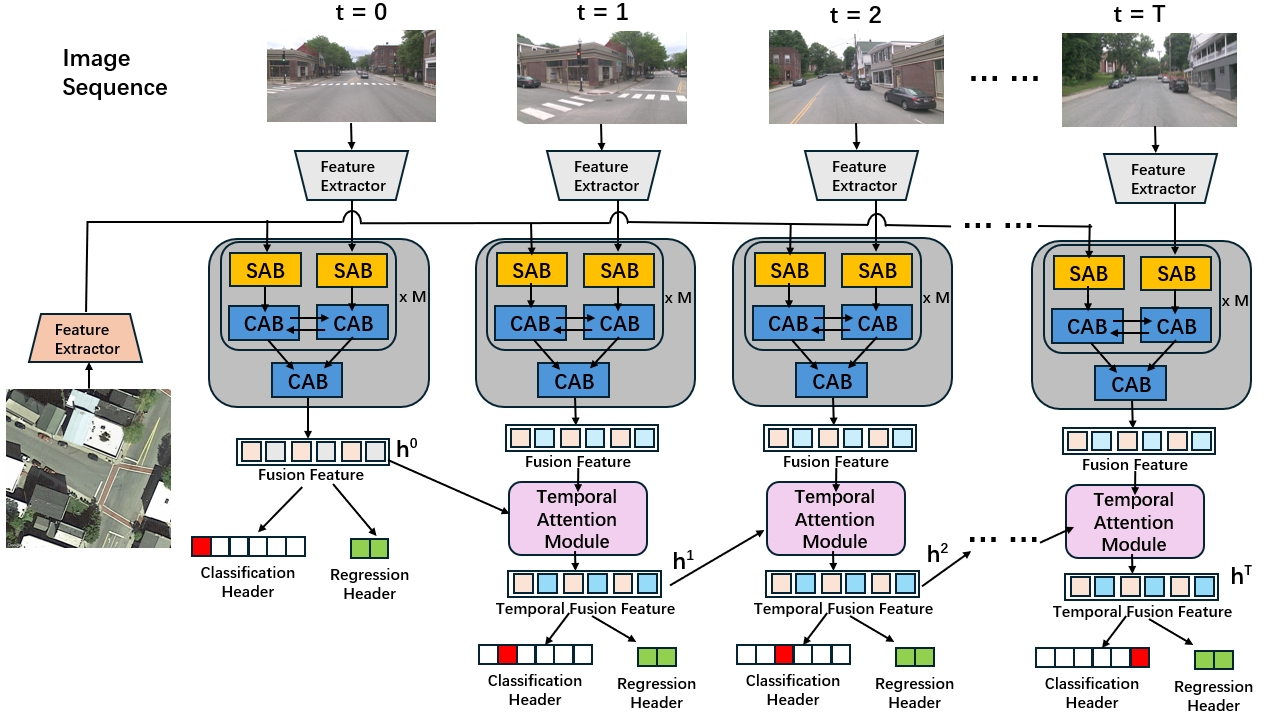}
    \caption{Overview of the proposed architecture. The feature extractors extract feature maps independently from two different views. These feature maps are then fed into self-attention blocks (SAB) and cross-attention blocks (CAB)  to generate fused features. The fusion features are further integrated with the hidden state $h^{t-1}$ from the previous time step within the temporal attention module. Finally, the obtained features are utilized for the prediction of locations.}
    \label{fig:framework}
\end{figure*}

\subsection{Cross-view Image Retrieval Using Image Sequence} \label{sec:cvsil}
Previous cross-view image retrieval methods focused on matching a single street-view image with satellite images in a reference dataset. ~\cite{shi2022cvlnet} and ~\cite{zhang2023cross} extended the one-to-one image retrieval work by using continuous street-view images to match a satellite image. ~\cite{shi2022cvlnet} proposed CVLNet to match a continuous frame of street-view images to a large satellite image for camera localization. The CVLNet transforms and projects sequential ground-view images into an overhead view and utilizes photo consistency among the projected images to form a global representation. To evaluate their model, they collected corresponding satellite images from Google Maps for the KITTI dataset~\cite{geiger2013vision} to construct the KITTI-CVL benchmark dataset. Similarly, ~\cite{zhang2023cross} employed a Transformer structure to fuse multiple consecutive ground-view images into a global descriptor for matching with a satellite image. Additionally, they also utilized street-view images captured by on-board cameras and satellite images sourced from Google Maps to build a large-scale cross-view image sequence geo-localization (CVIS) dataset. While these two studies are rooted in image retrieval and aim to match a street-view image sequence with a satellite image, our proposed novel task involves predicting the pixel coordinates of sequential street-view images in a satellite image. Theoretically, our new task is more useful for robot localization deployed in the real world. We evaluate the performance of our approach using the aforementioned two datasets, which will be detailed further in the Experiment Settings Section \ref{sec:experiment setting}.

\section{Methodology}
In this section, we present the proposed method for cross-view sequential image localization. We introduce a method that achieves state-of-the-art performance on cross-view fine-grained localization~\cite{yuan2024cross} as our baseline framework in Section \ref{sec:baseline}. Section \ref{sec:rnn} introduces the proposed \textit{Hidden State}, which retains information from the preceding time step. In Section \ref{sec:temporal-attention}, we present a novel Temporal Attention Module (TAM), which can integrate information from both the previous and current time steps. An overview of the proposed framework is presented in Fig.\ref{fig:framework}. Section \ref{sec:loss} introduces the training objectives.

\subsection{Baseline Framework} \label{sec:baseline}
Given that the proposed method in~\cite{yuan2024cross} yielded state-of-the-art outcomes in the cross-view fine-grained localization task, we adopted this method as our baseline framework. This framework primarily comprises local feature extractors, the cross-view feature fusion network, and prediction headers, which will be elaborated upon in the subsequent section.

\subsubsection{Local Feature Extractor}
Two modified ResNet50~\cite{he2016deep} neural networks are employed to extract features from the satellite image and ground-view image, respectively. Upon passing the ground-view image through the feature extractor, a feature map $F_g \in \mathbb{R}^ {C \times H_g \times W_g}$ is derived, where $H_g$ and $W_g$ are $1/8$ of the height and width of the original input ground-view image, and $C$ is the channel dimension. Similarly, the satellite image undergoes feature extraction, resulting in a feature map $F_s \in \mathbb{R}^ {C \times N \times N}$, where $N$ represents the dimensions of the extracted satellite image feature map. Subsequently, the resulting feature maps from two views are flattened, and a $1 \times 1$ convolution layer is utilized to reduce the channel dimension $C$ to $D$. The obtained features serve as input for the subsequent stage. 

\subsubsection{Cross-View Feature Fusion Network}
The cross-view feature fusion network primarily consists of self-attention blocks and cross-attention blocks, utilized to enhance and fuse image features, respectively. Features extracted from the ground-view and satellite images, $F_{g'} \in \mathbb{R}^ {D \times H_gW_g}$ and $F_{s'} \in \mathbb{R}^ {D \times N^2}$, are fed into a self-attention block (SAB) followed by a cross-attention block (CAB), respectively. In the cross-attention block, the features from one view (either ground-view or aerial view) provide Key and Value, while the features from the other view provide Query to conduct multi-head cross-attention operations. This process (an SAB followed by a CAB) repeats $M$ times and concludes with a CAB to derive the ultimate fusion feature $F_{f} \in \mathbb{R}^ {D \times N^2}$. 

\subsubsection{Prediction Headers} \label{sec:prediction_header}
Two prediction headers are constructed on the fusion features: a classification header and a regression header. Both prediction headers consist of three-layer perceptrons. The classification header produces an $N^2 \times 1$ vector, tasked with predicting the specific cell of the $N \times N$ grid where the street-view image is located. Meanwhile, the regression header is designed to predict the offsets associated with the grid selected by the classification header.

\subsection{Hidden State} \label{sec:rnn}
Recurrent Neural Networks (RNNs)~\cite{elman1990finding} are a class of neural networks specifically designed to process sequential data by maintaining an internal state or memory. This memory or internal state enables RNNs to capture temporal dependencies in sequential data, making them successful in tasks such as speech recognition~\cite{vinyals2012revisiting} and text generation~\cite{sutskever2011generating}. In the Cross-view sequential image localization task, we argue that two consecutive street-view images share overlapping semantic information. Consequently, leveraging information from the preceding time step is advantageous for accurately predicting the location at the current time step. Inspired by the idea of RNNs, we create a hidden state $h$ to retain information from the preceding time step. At time $t$, the satellite image $I^{t}_s$ and ground-view image $I^{t}_g$ are processed through local feature extractors and a cross-view feature fusion network (detailed in Section \ref{sec:baseline}), yielding the fusion feature $F^{t}_{f}$. Subsequently, the fusion feature $F^{t}_{f}$ and the hidden state $h^{t-1}$ from the previous time step are fed into our proposed Temporal Attention Module (to be elaborated upon in the subsequent section), resulting in the temporal information integrated feature $\hat{F}^{t}_{f}$. This derived feature is utilized to connect the prediction headers for location prediction. Concurrently, the derived feature $\hat{F}^{t}_{f}$ is stored as the hidden state $h^{t}$ at the current time step and will contribute to the temporal attention process at the next time step. At the first time step $t = 0$, unlike conventional RNNs, we do not initialize a hidden state with a zero vector. Instead, we directly store the fusion feature $F^{0}_{f}$ as the hidden state $h^{0}$ and utilize $F^{0}_{f}$ to predict the location of the first ground-view image. The aforementioned procedure can be described as follows:

\begin{equation}
    h^{t} = \begin{cases}
                F^{0}_{f} & t = 0  \\
                \mbox{TAM}(F^{t}_{f}, h^{t-1}) & t > 0
		   \end{cases}
\end{equation}

\begin{figure}
    \centering
    \includegraphics[width=1\linewidth]{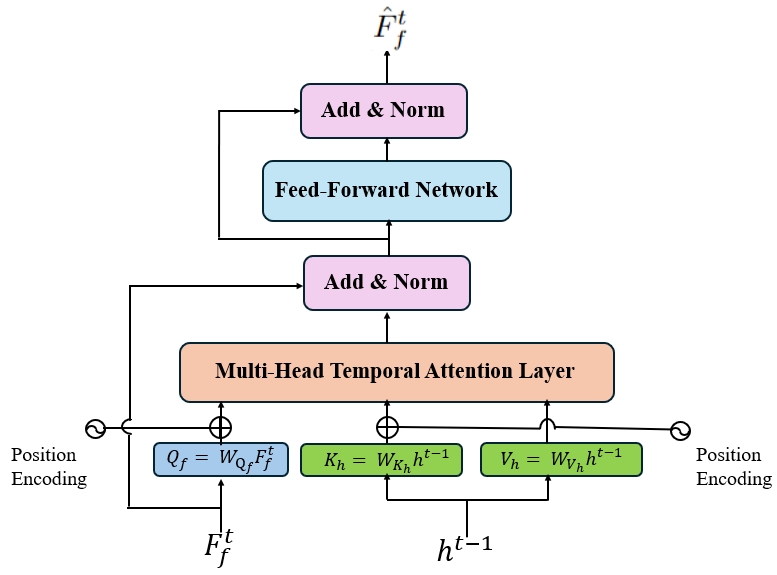}
    \caption{Illustration of Temporal Attention Module (TAM). 
The fusion feature $F^{t}_f$ undergoes projection to generate the Query $Q_f$, while the hidden state $h^{t-1}$ from the previous time step is projected to form the Key $K_h$ and Value $V_h$. The Position Encoding is applied to both $Q_f$ and $K_h$ before they are passed into the multi-head attention layers.}
    \label{fig:TAM}
\end{figure}

\subsection{Temporal Attention Module (TAM)} \label{sec:temporal-attention}
The proposed Temporal Attention Module aims to integrate temporal information from the previous time step into the current prediction process, shown in Fig.\ref{fig:TAM}. We achieve this by using the attention mechanism~\cite{vaswani2017attention}, where Q (query), K (key) and V (value) are fundamental components involved in the calculation of the attention mechanism. The objective is to capture the correlation between the current fusion feature $F^{t}_{f}$ and the hidden state $h^{t-1}$ through attention score computation. This correlation (the attention score) allows us to extract relevant information from the hidden state $h^{t-1}$, which is subsequently integrated with the fusion feature $F^{t}_{f}$ to derive a new feature $\hat{F}^{t}_{f}$. Hence, in this temporal attention process, the current fusion feature $F^{t}_{f}$ is projected to Q, while the hidden state $h^{t-1}$ is projected to both K and V. This projection process can be presented as:
\begin{equation}
    \begin{split}
        Q_f &= W_{Q_f} \, F^{t}_{f}\\
        K_h &= W_{K_h} \, h^{t-1}\\
        V_h &= W_{V_h} \, h^{t-1}
  \label{eq:qkv}
  \end{split}
\end{equation}
where $W_{Q_f}$, $W_{K_h}$ and $W_{V_h}$ are matrices to be learned. Following the approach in~\cite{carion2020end}, before multi-head attention layers, sinusoidal positional encodings $E_{pos}$ are added to $Q_f$ and $K_h$ respectively to maintain the positional information. Based on the attention score, relevant information $\tilde{F}_{tem}$ extracted from hidden state $h^{t-1}$ can be obtained by as follows:
\begin{equation}
     \tilde{F}_{tem} = \mbox{softmax}((Q_f + E_{pos})^T(K_h + E_{pos}))V_h
    \label{score}
\end{equation}
The relevant information $\tilde{F}_{tem}$ is added to the original fusion feature $F^{t}_{f}$ and undergoes processing through a feed-forward network (FFN), comprising two linear projection layers. Finally, the output feature $\hat{F}^{t}_{f}$ is obtained for position prediction and concurrently stored as the current hidden state $h^{t}$.

\subsection{Training Objective} \label{sec:loss}
Consistent with the approach outlined in~\cite{yuan2024cross}, we adopt the standard categorical cross-entropy loss function for training the classification header mentioned in \ref{sec:prediction_header}, which is defined as:

\begin{equation}
  \mathcal{L}_{cls} = - \sum_i^C y_i\log(\mbox{softmax}(p_i))
  \label{eq:cross-entropy}
\end{equation}
where $y_i$ represents the ground-truth label indicating the index of grid cells in an $N \times N$ grid and $p_i$ denotes the output of the classification header. The mean square error (MSE) is employed for training the offset regression header, which can be formulated as follows:  
\begin{equation}
    \mathcal{L}_{mse} = \frac{1}{N} \sum_{i=1}^N (y_i - \hat{y_i})^2
\end{equation}
where $y_i$ denotes the ground-truth offset values and $\hat{y_i}$ denotes the prediction offset values. 
In contrast to the approach proposed in~\cite{yuan2024cross}, our method receives a continuous sequence of street-view images as input. Consequently, we compute the average of the sequence loss as the objective loss for each iteration, which can be formulated as follows:

\begin{equation}
  % \mathcal{L}_{seq} = \frac{\sum^N_{n=1}(\lambda_{cls}\mathcal{L}^n_{cls} + \lambda_{mse}\mathcal{L}^n_{mse})}{N},
  \mathcal{L}_{seq} = \frac{1}{T} \sum_{t=1}^T \lambda_{cls}\mathcal{L}^t_{cls} + \lambda_{mse}\mathcal{L}^t_{mse}
  \label{eq:loss}
\end{equation}
where $T$ denotes the number of street-view images in a sequence.  $\lambda_{cls}$ and $\lambda_{mse}$ are the corresponding weight coefficients.

\section{Experimental Settings} \label{sec:experiment setting}

\subsection{Datasets}
We conduct experiments on two datasets, namely CVIS~\cite{zhang2023cross} and KITTI-CVL~\cite{shi2022cvlnet}. We choose the method proposed in~\cite{yuan2024cross} as our baseline and compare the performance on the cross-view sequential image localization task.

\subsubsection{CVIS Dataset}
The CVIS dataset consists of street-view images captured by a front-camera mounted on a vehicle in urban and suburban areas of the state of Vermont, USA. Each street-view image in the dataset has a resolution of 1920 $\times$ 1080 pixels and is accompanied by accurate GPS coordinates and heading direction information. Within a street-view image sequence, the distance between consecutive images is approximately 8 meters, while the distance between the first and last images does not exceed 50 meters. This constraint is implemented to ensure that a street-view image sequence can be covered within a satellite image. In total, the dataset comprises 38863 street-view image sequences, with each sequence containing an average of 7 images. Additionally, to represent more real-world scenarios, around 70\% of street-view images are collected from suburban areas and 30\% from urban areas.

Satellite images matching the number of street-view sequences are acquired using Google Maps. The GPS coordinates of the street-view image located in the middle of each sequence are selected and then randomly adjusted by a 5-meter offset to determine the coordinates of the corresponding satellite image. Each aerial image is captured at a zoom level of 20, yielding a resolution of 640 $\times$ 640 pixels. The ground resolution of satellite images is approximately 0.114 meters per pixel.

We split the CVIS dataset into training, validation and testing sets, comprising $24871$, $6217$, and $7773$ aerial-ground sequence pairs, respectively.

\begin{figure}
    \centering
    \includegraphics[width=1\linewidth]{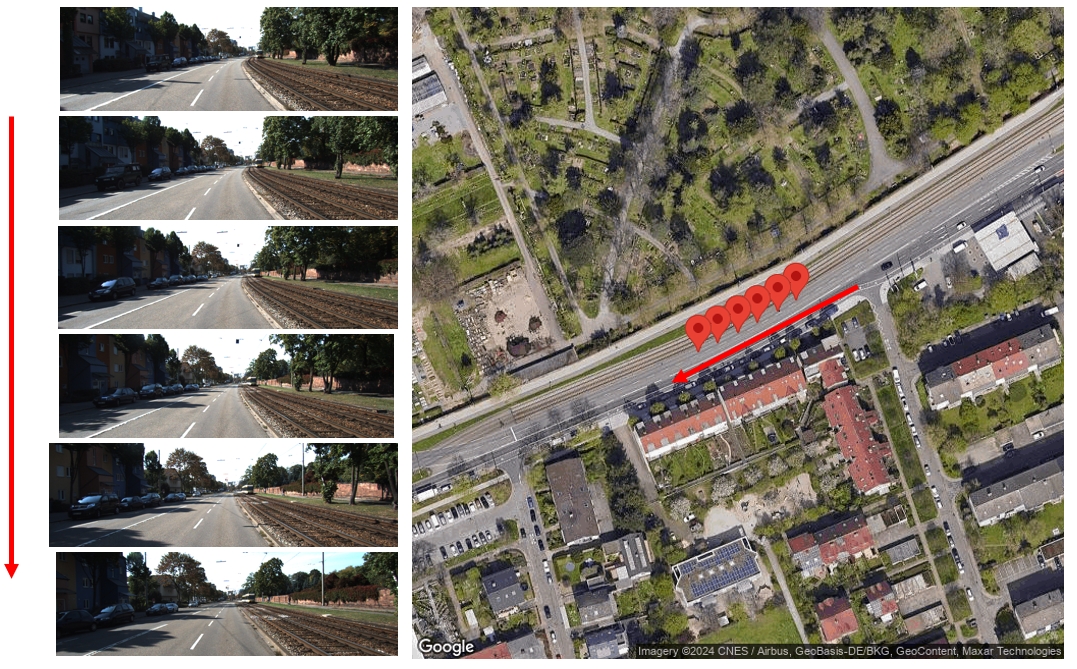}
    \caption{A sequence sample segmented from the KITTI-CVL dataset. Each red marker denotes the location of a street-view image. The red arrow indicates the direction of travel of the vehicle.}
    \label{fig:kitti}
\end{figure}

\subsubsection{KITTI-CVL dataset}
The KITTI dataset holds considerable significance in autonomous driving and computer vision research~\cite{geiger2013vision}. Shi \textit{et al.}~\cite{shi2022cvlnet} undertook the task of gathering satellite images that correspond to street-view images from Google Earth, thereby enriching the KITTI-CVL Dataset to investigate cross-view geo-localization challenges. Each satellite image in this dataset maintains a resolution of 1280 $\times$ 1280 pixels, with an associated ground resolution of approximately 20 centimetres per pixel.

\begin{figure*}
    \centering
    \includegraphics[width=1\linewidth]{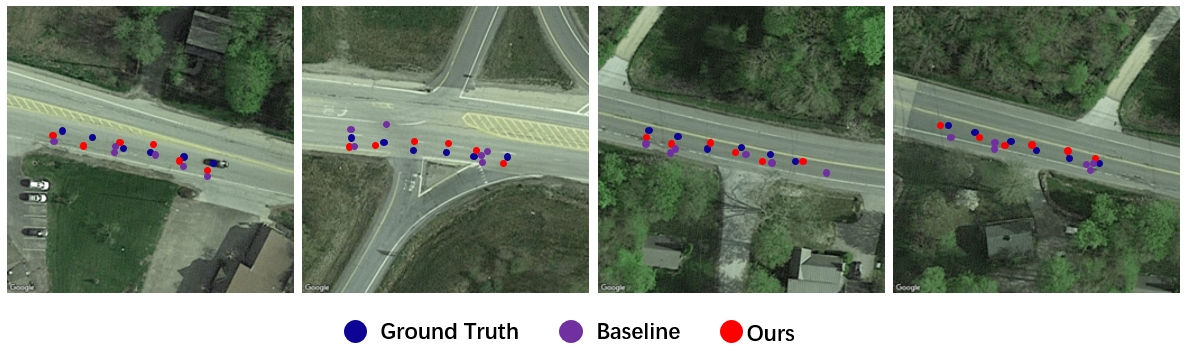}
    \caption{Qualitative localization result visualization on the CVIS~\cite{zhang2023cross} dataset. We use blue, purple and red points to denote the predicted consecutive street-view locations of ground truth, the baseline method and our proposed method. The figure illustrates that the baseline method tends to cluster multiple predicted locations in the same area, whereas our approach demonstrates a closer alignment with the actual ground truth locations.}
    \label{fig:visual}
\end{figure*}

In the original KITTI dataset, consecutive street-view images are highly similar with a small distance between them. To better suit practical applications of sequential image localization, we adopt a methodology akin to the CVIS dataset~\cite{zhang2023cross} by segmenting each complete video into multiple continuous street-view sequences. Specifically, we extract a frame every 8 meters from the entirety of the original video, generating a lengthy sequence denoted as $S = s_0, s_1, ..., s_X$. We commence by computing the distance between $s_0$ and $s_1$. If this distance falls below 50 meters, we proceed to calculate the distance between $s_0$ and $s_2$, iteratively checking until the distance between $s_0$ and $s_x$ exceeds 50 meters. Subsequently, $\bigr[s_0, s_x\bigr]$ constitutes a sequence. The next sequence starts from the midpoint of $\bigr[s_0, s_x\bigr]$. Upon deriving all sequences, those containing fewer than 6 street-view images are discarded, culminating in a final set of 1077 street-view image sequences. The satellite image corresponding to the midpoint location of each sequence is collected, yielding 1077 pairs of satellite images and corresponding street-view image sequences. A sample pair of ground-aerial images after being segmented from the KITTI-CVL dataset is shown in Fig.\ref{fig:kitti}. Similarly, all aerial-ground sequence pairs are subdivided into 862 pairs for training, 107 pairs for validation, and 108 pairs for testing.

\subsection{Implementation Details}
For the CVIS dataset, both satellite and street-view images are resized to dimensions of 256 $\times$ 256 and 270 $\times$ 480, respectively, before being fed into the model as input. Regarding the KITTI-CVL dataset, the satellite images undergo an initial center cropping to 640 $\times$ 640 before being resized to 256 $\times$ 256. Similarly, the street-view images are resized to 145 $\times$ 480 to prepare them for model input. We follow the settings in~\cite{yuan2024cross} to construct the baseline framework, where the channel dimension C $=$ 1024 of the feature maps obtained from the feature extractors, and will be reduced to D $=$ 256 through a 1 $\times$ 1 convolutional layer. The dimension N of the satellite image feature map and the classification grid N $\times$ N are both set to 32. The SAB followed by the CAB process will be repeated $M = 4$ times. We initialize the parameters of the local feature extractor using pre-trained weights from ResNet50~\cite{he2016deep}, trained on the ImageNet dataset~\cite{russakovsky2015imagenet}. The remaining parameters are initialized using Xavier initialization~\cite{glorot2010understanding}. For model optimization, we employ the AdamW optimizer~\cite{loshchilov2018decoupled}. Specifically, we set the learning rate for the feature extractors to $10^-5$, while the learning rate for other parameters is set to $10^-4$.

\subsubsection{Training Process}
Firstly, we initiate training for the baseline method~\cite{yuan2024cross}. This involves randomly selecting a street-view image from each sequence, pairing it with the corresponding satellite image covering the sequence, and feeding the pair into the model. The batch size is configured to $4$. During inference, the entire sequence is inputted, and the location of each street-view image will be predicted. Subsequently, we proceed to train our proposed new method. Here, we utilize the pre-trained feature extractors from the baseline method, keeping them frozen, to extract feature maps. The obtained feature maps will be fed into the subsequent training stage for training parameters of other components. Each input comprises a street-view image sequence paired with its corresponding satellite image, with a batch size of $1$. Each sequence contains 6 consecutive street-view images for training and testing. Regarding the KITTI-CVL dataset, the limited number of sequences obtained through segmentation raises concerns about overfitting during direct training. Therefore, we opt to fine-tune the model trained on the CVIS dataset using the KITTI-CVL data. 

\subsubsection{Evaluation Metrics}
In our evaluation process, we predict the locations of all street-view images within each sequence and subsequently compute the average distance error for the entire sequence, denoted as the sequence error. Upon collecting all sequence errors within the dataset, we derive both the mean and median values of all sequence errors. These two measures serve as evaluative metrics to assess the performance of the methods.

\begin{table}
\centering
\caption{Comparison between our method with single-image fine-grained localization methods on the CVIS dataset. The results are shown on the validation and test datasets. Best results are in bold.}
\label{CVIS_res}
\begin{center}
\begin{tabular}{|l||ll|ll|}
\hline
               & \multicolumn{2}{l|}{Validation}             & \multicolumn{2}{l|}{Test}                   \\ \hline
Localization error(m) & \multicolumn{1}{l|}{mean} & median & \multicolumn{1}{l|}{mean} & median \\ \hline
CVML~\cite{xia2022visual}           & \multicolumn{1}{l|}{15.99}         & 14.72           & \multicolumn{1}{l|}{16.37}         & 15.13           \\ \hline
Baseline~\cite{yuan2024cross}       & \multicolumn{1}{l|}{12.31}         & 11.84           & \multicolumn{1}{l|}{12.57}         & 12.15           \\ \hline
Ours           & \multicolumn{1}{l|}{\textbf{4.30}}          & \textbf{1.89}            & \multicolumn{1}{l|}{\textbf{3.29}}              &   \textbf{1.74}              \\ \hline
\end{tabular}
\end{center}
\end{table}

\section{Experiments Results}
\subsection{Results on CVIS Dataset}
The results from the comparison of cross-view sequential image localization on the CVIS dataset are summarized in Table \ref{CVIS_res}. Apart from the baseline method, another cross-view fine-grained localization method~\cite{xia2022visual} mentioned in Section \ref{sec:cvsil} is trained and tested on the CVIS dataset for comparison with our proposed method. As depicted in Table \ref{CVIS_res}, the performance of these two methods considering the location of a single street view is notably poor in predicting the location of continuous street-view images. This poor performance can be attributed to the similarity among street-view images in a sequence. Models that only consider a single image struggle to establish a robust correspondence between similar street views and different locations, leading to larger prediction distance errors.

The visualization results (see Fig.\ref{fig:visual}) further illustrate that some street-view images, which are relatively similar, are erroneously predicted to be in the same area by these models. In contrast, the newly proposed method incorporates a temporal attention module that compares and integrates information from the previous time step with that of the current time step. This mechanism enhances the accuracy of the model in predicting sequential locations. Compared to the baseline method, our proposed method achieves a significant improvement in both mean distance error and median distance error on the test dataset, with enhancements of 73.8\% and 85.7\%, respectively.

\subsection{Generalization in Unknown Area across Times}
The generalization ability of our method is assessed on unknown regions and different dates using the KITTI-CVL dataset. The street-view images in the KITTI dataset utilized for training and testing were collected across various times and areas. The experimental results are presented in Table \ref{KITTI_res}.

\begin{table}
\centering
\caption{Comparison between our method with the baseline method on the KITTI-CVL dataset. The results are shown on the validation and test datasets. Best results are in bold.}
\label{KITTI_res}
\begin{center}
\begin{tabular}{|l||ll|ll|}
\hline
               & \multicolumn{2}{l|}{Validation}             & \multicolumn{2}{l|}{Test}                   \\ \hline
Localization error(m) & \multicolumn{1}{l|}{mean} & median & \multicolumn{1}{l|}{mean} & median \\ \hline
Baseline~\cite{yuan2024cross}           & \multicolumn{1}{l|}{13.56}         & 13.17           & \multicolumn{1}{l|}{13.91}         &13.47            \\ \hline
Ours       & \multicolumn{1}{l|}{11.37}         &    10.53        & \multicolumn{1}{l|}{12.45}         &   11.91         \\ \hline
Ours + Fine-tuning           & \multicolumn{1}{l|}{\textbf{3.00}}          & \textbf{1.70}            & \multicolumn{1}{l|}{\textbf{3.07}}              &       \textbf{2.10}          \\ \hline
\end{tabular}
\end{center}
\end{table}

In Table \ref{KITTI_res}, the first two rows display the outcomes of the baseline method and our new method directly trained on the KITTI-CVL dataset. It is evident that our new method outperforms the baseline method. Remarkably, when fine-tuning the model trained on the CVIS dataset with the KITTI-CVL dataset, there is a significant improvement in prediction accuracy. This indicates that our method exhibits strong generalization ability across different location areas and dates. Moreover, our model achieves commendable performance across diverse data domains through fine-tuning with a small amount of data, highlighting its generalization capability across different domains.

\section{Conclusion and Future Work}
In this work, we expand upon the cross-view fine-grained localization problem to address the cross-view sequential image localization challenge, which aligns more closely with real-world applications. To tackle this task, we introduce a novel temporal attention module designed to integrate information from both the preceding and current time steps, thereby enhancing the accuracy of predicting the position of street-view images at the current step. Furthermore, we partition the KITTI-CVL dataset into shorter street-view image sequences, facilitating the evaluation of cross-view sequence image localization. Through extensive experimentation, our proposed method demonstrates notable effectiveness in predicting sequential street-view image locations. Additionally, it exhibits superior generalization capabilities in unknown areas across different times. Future work will incorporate motion prior information to enhance the accuracy of cross-view sequential image localization in large-scale areas.

% \input{introduction}
% \input{mathstuff}
% \input{related_work}
% % \input{PRELIMINARIES}
% \input{Method}
% \input{expriment_setting2}
% \input{results}
% \input{conclusion}

% \addtolength{\textheight}{-12cm}   

%\section*{ACKNOWLEDGMENT}
%The preferred spelling of the word ÒacknowledgmentÓ in America is without an ÒeÓ after the ÒgÓ. Avoid the stilted expression, ÒOne of us (R. B. G.) thanks . . .Ó  Instead, try ÒR. B. G. thanksÓ. Put sponsor acknowledgments in the unnumbered footnote on the first page.
%%%%%%%%%%%%%%%%%%%%%%%%%%%%%%%%%%%%%%%%%%%%%%%%%%%%%%%%%%%%%%%%%%%%%%%%%%%%%%%%
\bibliographystyle{IEEEtran}
\bibliography{ref.bib}

\begin{thebibliography}{10}
\providecommand{\url}[1]{#1}
\csname url@samestyle\endcsname
\providecommand{\newblock}{\relax}
\providecommand{\bibinfo}[2]{#2}
\providecommand{\BIBentrySTDinterwordspacing}{\spaceskip=0pt\relax}
\providecommand{\BIBentryALTinterwordstretchfactor}{4}
\providecommand{\BIBentryALTinterwordspacing}{\spaceskip=\fontdimen2\font plus
\BIBentryALTinterwordstretchfactor\fontdimen3\font minus \fontdimen4\font\relax}
\providecommand{\BIBforeignlanguage}[2]{{%
\expandafter\ifx\csname l@#1\endcsname\relax
\typeout{** WARNING: IEEEtran.bst: No hyphenation pattern has been}%
\typeout{** loaded for the language `#1'. Using the pattern for}%
\typeout{** the default language instead.}%
\else
\language=\csname l@#1\endcsname
\fi
#2}}
\providecommand{\BIBdecl}{\relax}
\BIBdecl

\bibitem{hu2018cvm}
S.~Hu, M.~Feng, R.~M. Nguyen, and G.~H. Lee, ``Cvm-net: Cross-view matching network for image-based ground-to-aerial geo-localization,'' in \emph{Proceedings of the IEEE Conference on Computer Vision and Pattern Recognition}, 2018, pp. 7258--7267.

\bibitem{shi2019spatial}
Y.~Shi, L.~Liu, X.~Yu, and H.~Li, ``Spatial-aware feature aggregation for image based cross-view geo-localization,'' \emph{Advances in Neural Information Processing Systems}, vol.~32, 2019.

\bibitem{regmi2019bridging}
K.~Regmi and M.~Shah, ``Bridging the domain gap for ground-to-aerial image matching,'' in \emph{Proceedings of the IEEE/CVF International Conference on Computer Vision}, 2019, pp. 470--479.

\bibitem{liu2019lending}
L.~Liu and H.~Li, ``Lending orientation to neural networks for cross-view geo-localization,'' in \emph{Proceedings of the IEEE/CVF conference on computer vision and pattern recognition}, 2019, pp. 5624--5633.

\bibitem{shi2020optimal}
Y.~Shi, X.~Yu, L.~Liu, T.~Zhang, and H.~Li, ``Optimal feature transport for cross-view image geo-localization,'' in \emph{Proceedings of the AAAI Conference on Artificial Intelligence}, vol.~34, no.~07, 2020, pp. 11\,990--11\,997.

\bibitem{zhu2022transgeo}
S.~Zhu, M.~Shah, and C.~Chen, ``Transgeo: Transformer is all you need for cross-view image geo-localization,'' in \emph{Proceedings of the IEEE/CVF Conference on Computer Vision and Pattern Recognition}, 2022, pp. 1162--1171.

\bibitem{shi2020looking}
Y.~Shi, X.~Yu, D.~Campbell, and H.~Li, ``Where am i looking at? joint location and orientation estimation by cross-view matching,'' in \emph{Proceedings of the IEEE/CVF Conference on Computer Vision and Pattern Recognition}, 2020, pp. 4064--4072.

\bibitem{cai2019ground}
S.~Cai, Y.~Guo, S.~Khan, J.~Hu, and G.~Wen, ``Ground-to-aerial image geo-localization with a hard exemplar reweighting triplet loss,'' in \emph{Proceedings of the IEEE/CVF International Conference on Computer Vision}, 2019, pp. 8391--8400.

\bibitem{yang2021cross}
H.~Yang, X.~Lu, and Y.~Zhu, ``Cross-view geo-localization with layer-to-layer transformer,'' \emph{Advances in Neural Information Processing Systems}, vol.~34, pp. 29\,009--29\,020, 2021.

\bibitem{yuan2024cross}
D.~Yuan, F.~Maire, and F.~Dayoub, ``Cross-attention between satellite and ground views for enhanced fine-grained robot geo-localization,'' in \emph{Proceedings of the IEEE/CVF Winter Conference on Applications of Computer Vision}, 2024, pp. 1249--1256.

\bibitem{xia2022visual}
Z.~Xia, O.~Booij, M.~Manfredi, and J.~F. Kooij, ``Visual cross-view metric localization with dense uncertainty estimates,'' in \emph{European Conference on Computer Vision}.\hskip 1em plus 0.5em minus 0.4em\relax Springer, 2022, pp. 90--106.

\bibitem{zhu2021vigor}
S.~Zhu, T.~Yang, and C.~Chen, ``Vigor: Cross-view image geo-localization beyond one-to-one retrieval,'' in \emph{Proceedings of the IEEE/CVF Conference on Computer Vision and Pattern Recognition}, 2021, pp. 3640--3649.

\bibitem{zhang2023cross}
X.~Zhang, W.~Sultani, and S.~Wshah, ``Cross-view image sequence geo-localization,'' in \emph{Proceedings of the IEEE/CVF Winter Conference on Applications of Computer Vision}, 2023, pp. 2914--2923.

\bibitem{shi2022cvlnet}
Y.~Shi, X.~Yu, S.~Wang, and H.~Li, ``Cvlnet: Cross-view feature correspondence learning for video-based camera localization,'' in \emph{Proceedings of the Asian Conference on Computer Vision}, 2022, pp. 652--669.

\bibitem{arandjelovic2016netvlad}
R.~Arandjelovic, P.~Gronat, A.~Torii, T.~Pajdla, and J.~Sivic, ``Netvlad: Cnn architecture for weakly supervised place recognition,'' in \emph{Proceedings of the IEEE conference on computer vision and pattern recognition}, 2016, pp. 5297--5307.

\bibitem{isola2017image}
P.~Isola, J.-Y. Zhu, T.~Zhou, and A.~A. Efros, ``Image-to-image translation with conditional adversarial networks,'' in \emph{Proceedings of the IEEE conference on computer vision and pattern recognition}, 2017, pp. 1125--1134.

\bibitem{toker2021coming}
A.~Toker, Q.~Zhou, M.~Maximov, and L.~Leal-Taix{\'e}, ``Coming down to earth: Satellite-to-street view synthesis for geo-localization,'' in \emph{Proceedings of the IEEE/CVF Conference on Computer Vision and Pattern Recognition}, 2021, pp. 6488--6497.

\bibitem{vaswani2017attention}
A.~Vaswani, N.~Shazeer, N.~Parmar, J.~Uszkoreit, L.~Jones, A.~N. Gomez, {\L}.~Kaiser, and I.~Polosukhin, ``Attention is all you need,'' \emph{Advances in neural information processing systems}, vol.~30, 2017.

\bibitem{dosovitskiy2020image}
A.~Dosovitskiy, L.~Beyer, A.~Kolesnikov, D.~Weissenborn, X.~Zhai, T.~Unterthiner, M.~Dehghani, M.~Minderer, G.~Heigold, S.~Gelly \emph{et~al.}, ``An image is worth 16x16 words: Transformers for image recognition at scale,'' in \emph{International Conference on Learning Representations}, 2020.

\bibitem{carion2020end}
N.~Carion, F.~Massa, G.~Synnaeve, N.~Usunier, A.~Kirillov, and S.~Zagoruyko, ``End-to-end object detection with transformers,'' in \emph{European conference on computer vision}.\hskip 1em plus 0.5em minus 0.4em\relax Springer, 2020, pp. 213--229.

\bibitem{geiger2013vision}
A.~Geiger, P.~Lenz, C.~Stiller, and R.~Urtasun, ``Vision meets robotics: The kitti dataset,'' \emph{The International Journal of Robotics Research}, vol.~32, no.~11, pp. 1231--1237, 2013.

\bibitem{he2016deep}
K.~He, X.~Zhang, S.~Ren, and J.~Sun, ``Deep residual learning for image recognition,'' in \emph{Proceedings of the IEEE conference on computer vision and pattern recognition}, 2016, pp. 770--778.

\bibitem{elman1990finding}
J.~L. Elman, ``Finding structure in time,'' \emph{Cognitive science}, vol.~14, no.~2, pp. 179--211, 1990.

\bibitem{vinyals2012revisiting}
O.~Vinyals, S.~V. Ravuri, and D.~Povey, ``Revisiting recurrent neural networks for robust asr,'' in \emph{2012 IEEE international conference on acoustics, speech and signal processing (ICASSP)}.\hskip 1em plus 0.5em minus 0.4em\relax IEEE, 2012, pp. 4085--4088.

\bibitem{sutskever2011generating}
I.~Sutskever, J.~Martens, and G.~E. Hinton, ``Generating text with recurrent neural networks,'' in \emph{Proceedings of the 28th international conference on machine learning (ICML-11)}, 2011, pp. 1017--1024.

\bibitem{russakovsky2015imagenet}
O.~Russakovsky, J.~Deng, H.~Su, J.~Krause, S.~Satheesh, S.~Ma, Z.~Huang, A.~Karpathy, A.~Khosla, M.~Bernstein \emph{et~al.}, ``Imagenet large scale visual recognition challenge,'' \emph{International journal of computer vision}, vol. 115, pp. 211--252, 2015.

\bibitem{glorot2010understanding}
X.~Glorot and Y.~Bengio, ``Understanding the difficulty of training deep feedforward neural networks,'' in \emph{Proceedings of the thirteenth international conference on artificial intelligence and statistics}.\hskip 1em plus 0.5em minus 0.4em\relax JMLR Workshop and Conference Proceedings, 2010, pp. 249--256.

\bibitem{loshchilov2018decoupled}
I.~Loshchilov and F.~Hutter, ``Decoupled weight decay regularization,'' in \emph{International Conference on Learning Representations}, 2018.

\end{thebibliography}
\end{document}